\documentclass[letterpaper, 10 pt, conference]{ieeeconf}
\IEEEoverridecommandlockouts
\usepackage{multicol}
\usepackage[bookmarks=true]{hyperref}

\usepackage{amssymb}
\usepackage{amsmath}
\usepackage{float}
\usepackage[noadjust]{cite}
\usepackage{graphicx}
\usepackage{hyperref}
\usepackage{mathtools}

\usepackage{filecontents}
\usepackage[font=footnotesize,labelfont=footnotesize]{subcaption}
\usepackage[font=footnotesize,labelfont=footnotesize]{caption}
\usepackage{subcaption}
\usepackage{caption}
\usepackage{comment}
\usepackage{authblk}
\usepackage{multirow}
\usepackage{environ}

\usepackage{amsthm}
\usepackage{algorithm, algorithmicx}
\usepackage[noend]{algpseudocode}
\usepackage[english]{babel}
\usepackage{xcolor}
\def\BibTeX{{\rm B\kern-.05em{\sc i\kern-.025em b}\kern-.08em
    T\kern-.1667em\lower.7ex\hbox{E}\kern-.125emX}}
    
\graphicspath{ {img/} }

\begin{document}

\title{PuzzleBots: Physical Coupling of Robot Swarms}
\author{Sha Yi \and Zeynep Temel \and Katia Sycara
\thanks{The authors are with the Robotics Institute, Carnegie Mellon University, Pittsburgh, PA 15213, USA. Email: {\tt\small \{shayi, ztemel, katia\}@cs.cmu.edu}.}
\thanks{This work was funded by AFOSR award FA9550-18-1-0097 and AFRL/AFOSR award FA9550-18-1-0251.}
\thanks{Hardware and software implementation available at: {\tt\url{https://github.com/ZoomLab-CMU/puzzlebot}}. Video available: {\tt\url{https://www.youtube.com/watch?v=QP3eMZXLSw4}}}
}

\maketitle

\begin{abstract}
Robot swarms have been shown to improve the ability of individual robots by inter-robot collaboration. In this paper, we present the PuzzleBots - a low-cost robotic swarm system where robots can physically couple with each other to form functional structures with minimum energy consumption while maintaining individual mobility to navigate within the environment. Each robot has knobs and holes along the sides of its body so that the robots can couple by inserting the knobs into the holes. We present the characterization of knob design and the result of gap-crossing behavior with up to nine robots. We show with hardware experiments that the robots are able to couple with each other to cross gaps and decouple to perform individual tasks. We anticipate the PuzzleBots will be useful in unstructured environments as individuals and coupled systems in real-world applications.
\end{abstract}

\section{Introduction}
Collaborative swarm behaviors have been widely observed in nature. Ants have shown the ability to create functional structures  like bridges by joining together and collaboratively performing tasks in a complex environment \cite{reid2015army, graham2017optimal}. Robots face similar challenges when they operate in uncertain environmental conditions, for instance gaps and holes may block the navigation of robots, particularly those having small characteristic lengths. In such scenarios, the ability of robots can be extended using physical coupling to form a functional swarm system and continue performing the designated tasks.

Large groups of robots have shown to improve the efficiency and robustness of task performances \cite{luo2020behavior, olfati2007consensus, pickem2017robotarium}.  It has also been shown that physically coupled structures of modular robots \cite{wright2007design, davey2012Emulating} can navigate in confined spaces and go over small gaps. 
In modular robots whose modules are initially coupled \cite{wright2007design, yim2000polybot}, each module has limited capability to navigate around the environment. Compared to a multi-robot system where there is no physical connection between robots, coupled modular robots are also less robust to module failure, meaning that if one of the modules fails during execution, the entire system may be at risk. Most modular robots address this issue by adding complicated coupling mechanisms \cite{romanishin2013Mblocks, tosun2016Design} that consumes additional power, which is already limited on a small module. 

Based on the above limitations, our goal with the PuzzleBots system is to build a robotic swarms system where 1) individual robots can dynamically couple and decouple with each other, 2) the coupling mechanism consumes minimum energy so that the main tasks of each robot are not influenced, 3) there is sufficient mobility and controllability of each individual robot to navigate within the environment, and 4) the fabrication of the robot is easy and cost is low so that it is possible to manufacture robots in larger quantity as a swarm system, even outside of the lab environment, i.e. at the task site. We will study the coupling mechanism based on passive connections - no additional components or power involved to perform the coupling behavior. To the best of our knowledge, this is the first work that utilizes passive connections, instead of active connections \cite{romanishin2013Mblocks, tosun2016Design, romanishin20153D, davey2012Emulating}, to form functional structures without sacrificing the mobility of each robot
and minimizing energy consumption of coupling mechanism.

\begin{figure}
    \centering
    \includegraphics[width=0.5\textwidth]{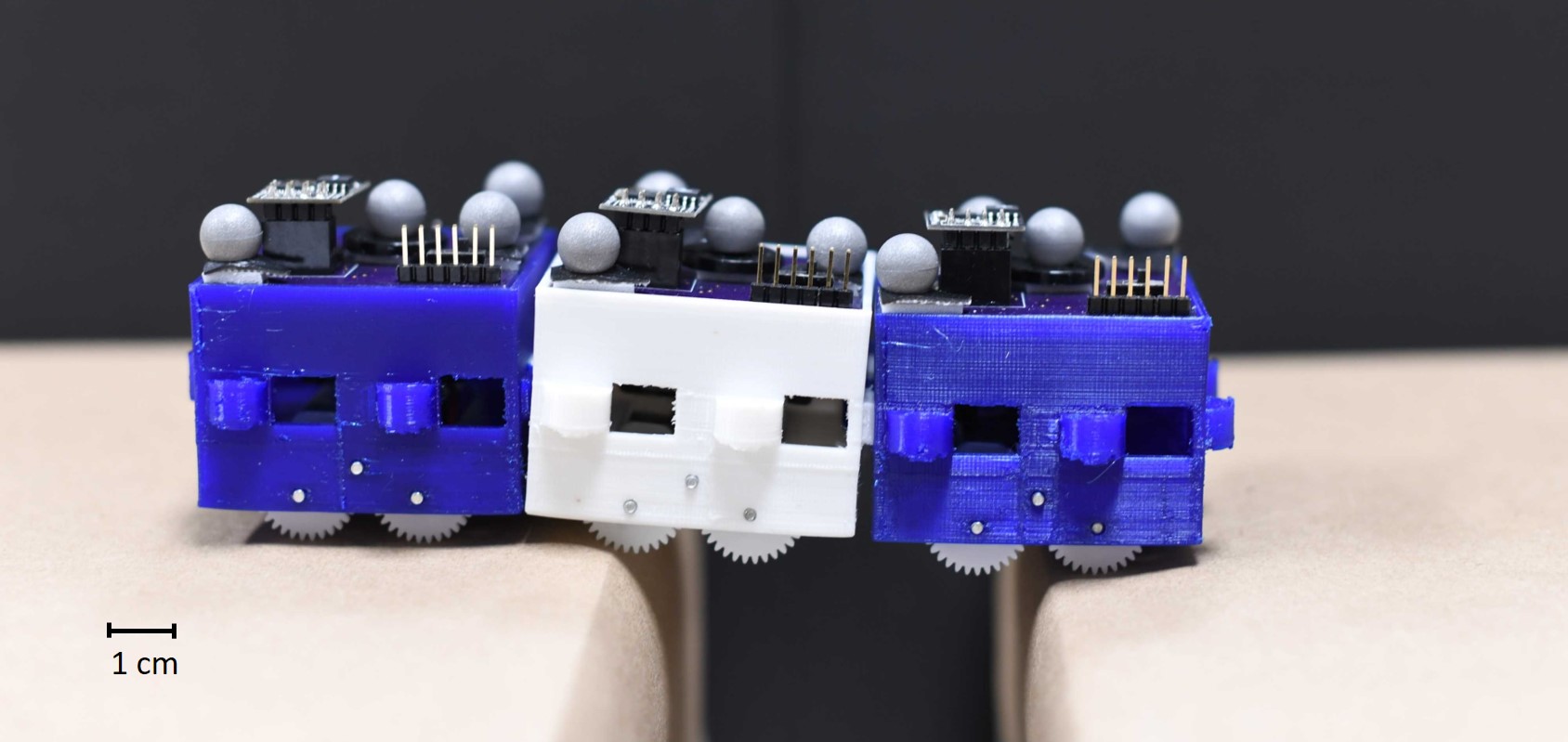}
    \caption{Three robots collaborate to cross a gap between two platforms.}
    \label{fig:cross_gap_demo}
    \vspace{-10px}
\end{figure}

The outline of the paper is as follows. In Section~\ref{sec:related_works}, we give an overview of the related works in robotic swarm 
systems and modular robots. Section~\ref{sec:methods} presents the detailed methods of our hardware and software platforms, including the mechanical design of the coupling mechanism, electronics of PuzzleBot, and the controller of the robots. In Section~\ref{sec:experiments}, we present experiments of characterizing the knobs for coupling and the results of the coupled-system performing gap-crossing motility. Finally, in Section~\ref{sec:conclusion}, we conclude our results and discuss future work.

\section{Related Works}\label{sec:related_works}
Robotics systems that involve interaction between multiple robots are able to demonstrate broad, dynamic, and collective behaviors \cite{yang2018grand, desai2001modeling, rubenstein2014programmable, rubenstein2012kilobot, pickem2017robotarium, arpino2018using}. Control algorithms have been extensively studied in swarm systems where robots do not physically interact with each other \cite{borrmann2015control, pickem2017robotarium, rubenstein2014programmable, arpino2018using, olfati2007consensus}. Robotic systems that actively leverage physical connections between robots lie mostly within the domain of modular self-reconfigurable robots \cite{yim2007modular}. Modular robots have shown exceptional performance in their flexibility and versatility to self-reconfigure for different tasks \cite{yim2000polybot, rollinson2016pipe}. Modular robots can be classified into two groups by the connection types between the modules. In the first group, robotic modules are connected throughout their execution, and the research focuses on controlling the configuration of the modules with respect to each other \cite{wright2007design}. Since the connection is not detachable, the flexibility of this group of robots is limited to a single connected component, and they are not resilient to module failures during execution. The other group includes the ATRON \cite{jorgensen2004modular}, M-TRAN III \cite{kurokawa2008distributed}, SlimeBot \cite{shimizu2009amoeboid}, Lily \cite{haghighat2015lily}, M-blocks \cite{romanishin20153D}, SMORES \cite{tosun2016Design}, FreeBot\cite{liang2020freebot}, and Swarm-bot \cite{gross2006autonomous} in which robotic modules can couple and decouple during execution. 
The Lily robots rely on external actuation from the fluid to connect with each other. ATRON, M-TRAN III and M-blocks do not require external actuation and flexibly connect with other modules using magnetic forces. Each module with the ATRON and M-TRAN III systems has limited mobility on their own. While the M-blocks modules can move by flipping along one of their axes, mobility of individual modules are limited compared with a standard wheeled robots, for example \cite{pickem2017robotarium}. 
SlimeBot, SMORES, FreeBot, and Swarm-bot modules are able to move around the environment independently. The SlimeBots connections are loosely couple and the main purpose is to communicate between robots, thus cannot bear any load. SMORES and FreeBot utilize magnetic forces for connection. The SMORES connection can bear the load of six robots. However, if the modules are misaligned, it can only support the weight of one module. Electromagnetic connections may also consumes high power \cite{liang2020freebot} with high loads. Permanent magnets do not consumer power, but will require additional power when separating the magnets.
Swarm-bot has independent grippers and complex connection mechanism, thus may not be able to carry load multiple times of its weight. Our proposed system aims to overcome the challenges of the above examples; it consists of individual low-cost robots that can perform as a swarm system while having the capability to physically couple with each other. 

\section{Methods}\label{sec:methods}

The goal of PuzzleBots robotic swarm
system is to demonstrate inter-robot collaboration by physically coupling with each other to form flexible, functional structures using a large number of robots. Therefore, the design considerations are as follows:
\begin{itemize}
\item Each robot is equipped with a coupling mechanism that enables dynamic coupling and decoupling behaviors with multiple robots.
\item To accommodate each robot's task performance, the coupling mechanism should consume minimum energy during execution.
\item Sufficient mobility and controllability are required so that each robot can navigate and complete tasks in the environment on its own.
\item To make it financially viable to build a system with a large number of robots, the cost of each robot should be kept as low as possible.
\end{itemize}
This section will introduce our first design of the PuzzleBots prototype that fulfills the requirements mentioned above.

\subsection{Robot Design}\label{sec:robot_design}

\begin{figure*}
\centering
\begin{subfigure}{0.32\textwidth}
\includegraphics[width=\textwidth]{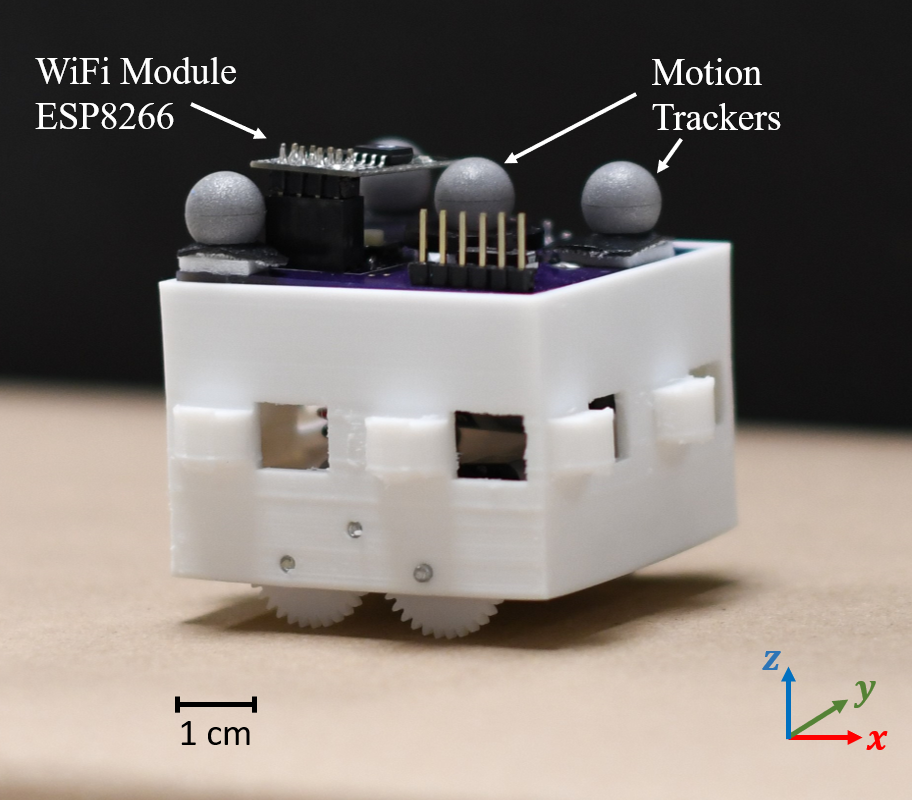}
\caption{}
\label{fig:robot_body}
\end{subfigure}
\begin{subfigure}{0.3\textwidth}
\includegraphics[width=\textwidth]{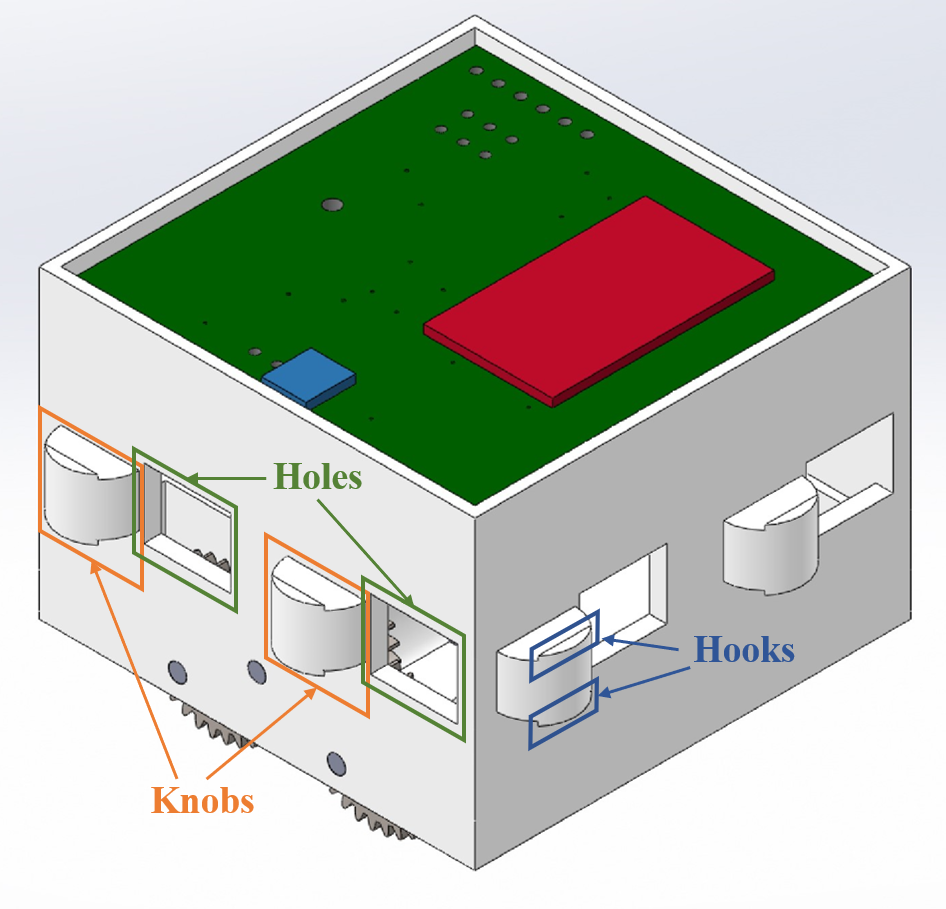}
\caption{}
\label{fig:assem_top45}
\end{subfigure}
\begin{subfigure}{0.325\textwidth}
\includegraphics[width=\textwidth]{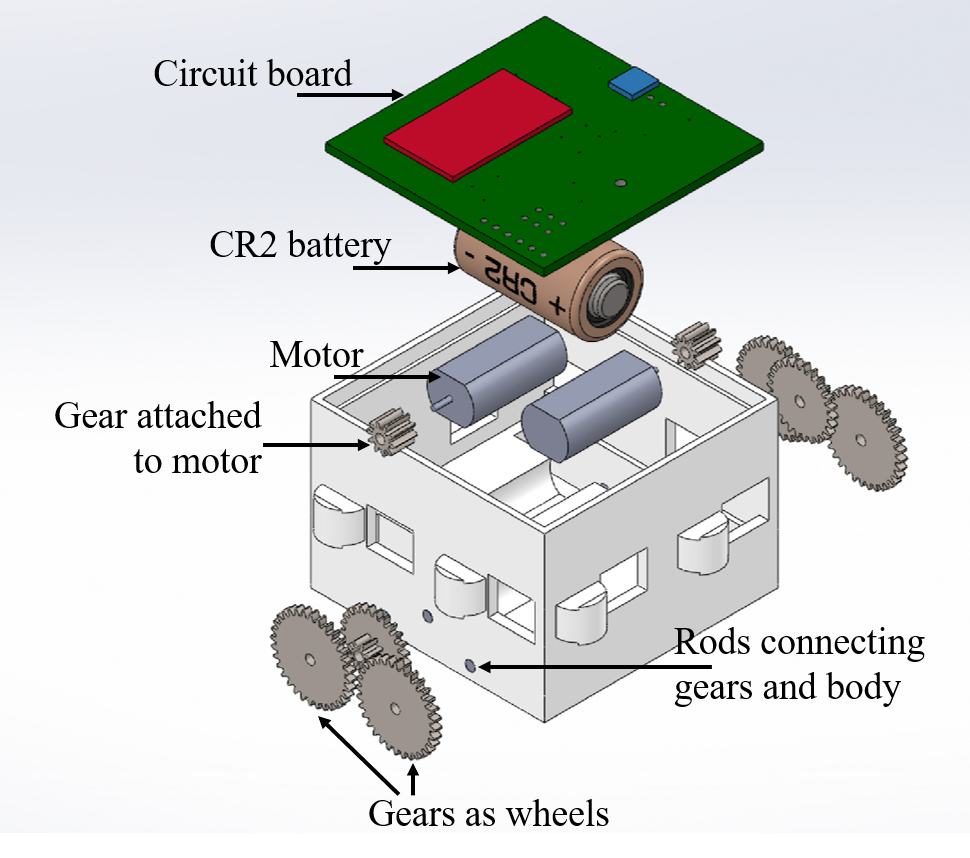}
\caption{}
\label{fig:robot_explode}
\end{subfigure}
\setlength{\belowcaptionskip}{-10pt}
\caption{(a) A PuzzleBot with motion trackers in Vicon system. The $xyz$ axis of the robot body frame points front, left, and up, correspondingly. (b) Design of the assembled robot. Each side of the robot body consists of two knobs and holes. There are hooks on top and bottom of each knob. (c) An exploded view of the PuzzleBot with mechanical and electrical parts.
}
\label{fig:robot_design}
\end{figure*}

Figure~\ref{fig:robot_body} shows our first generation of PuzzleBots.

Each robot weighs 62 g, including battery and four motion trackers to be used in the Vicon motion-tracking System\footnote{\tt{\url{https://www.vicon.com/}}}. A robot can carry a weight of 400 g, more than six times its own weight. Robots are equipped with on-board power, actuation, communication, and computation components. The coupling mechanism is inspired by the jigsaw puzzle. The body of the robot is 3D printed with thermoplastic polyurethane (TPU), consisting of eight knobs and holes equally distributed along the four sides of the robot, as shown in Figure~\ref{fig:assem_top45}. There are two hooks on the outer side of  each knob, one on the top and one on the bottom. Detailed explanation about the working principle will be provided in Section~\ref{sec:connection}. 
Each robot is 50 mm in width, 50 mm in depth, and 35 mm in height, excluding the knobs.
The mechanical and electrical components are placed inside the robot as seen in Figure~\ref{fig:robot_explode}, where each component will be explained in Section~\ref{sec:electronics} and Section~\ref{sec:mech_and_controls}.

As cost and time play vital roles in building systems with a large number of robots. We limit the cost by choosing commercially available parts. Each robot costs around US\$33.8, including - printing cost of the body (US\$3), CR2 battery (US\$2.45), two DC motors (US\$3.64 each), ESP8266 WiFi Module (US\$6.95), gears and rods (US\$1.6), double-sided circuit board (US\$5.7), and all other on-board electronics (approximate US\$6.8). Prices of small parts are computed based on purchasing quantity of 10-15 since purchasing in bulk may reduce the price. The time for 3D printing a robot chassis takes 4 hours and the assembly time for each robot takes approximately 30 minutes.

\subsection{Coupling Mechanism Design}\label{sec:connection}
\begin{figure*}
\begin{subfigure}{0.28\textwidth}
\includegraphics[width=\textwidth]{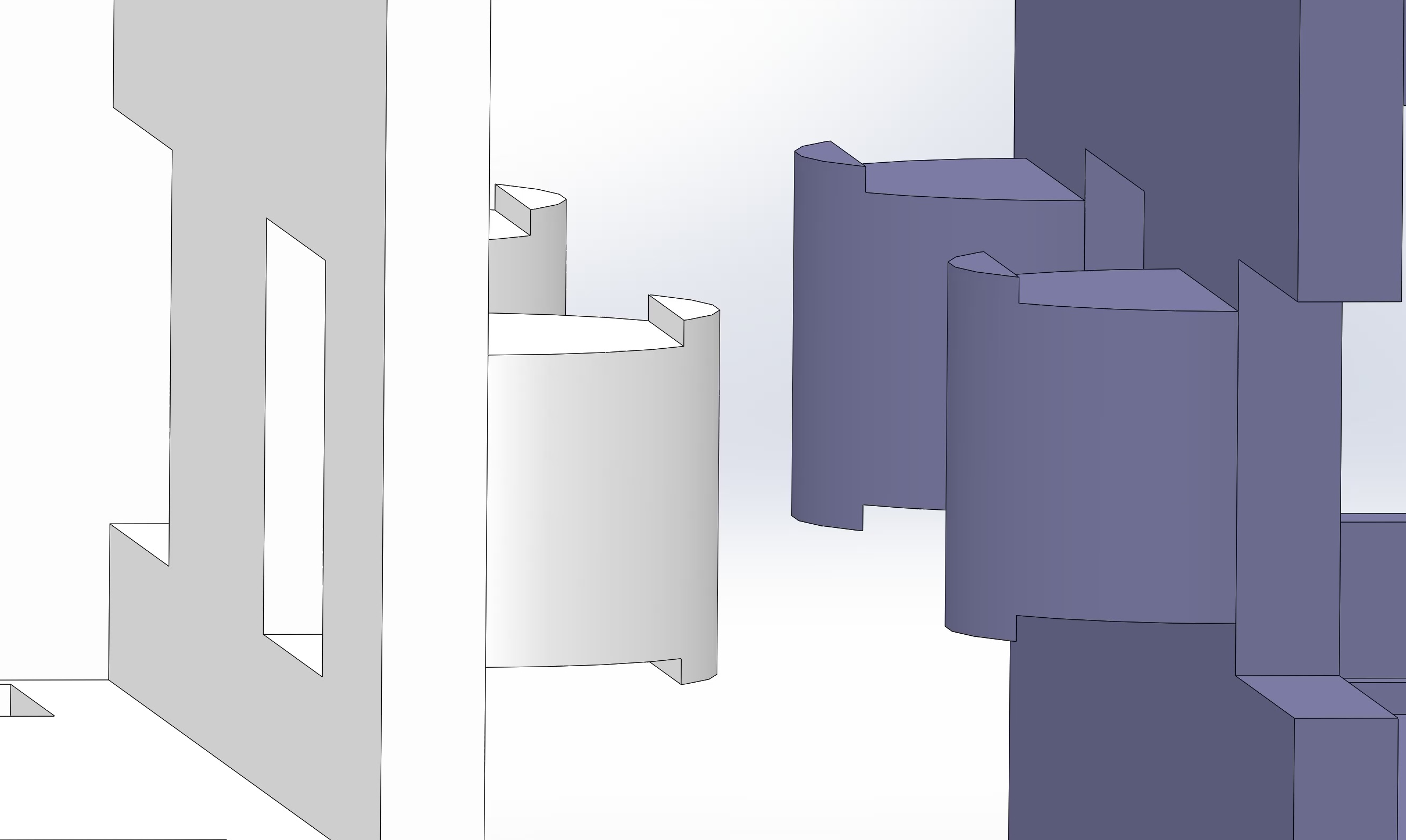}
\caption{}
\label{fig:connect_not_insert}
\end{subfigure}\hfill
\begin{subfigure}{0.28\textwidth}
\includegraphics[width=\textwidth]{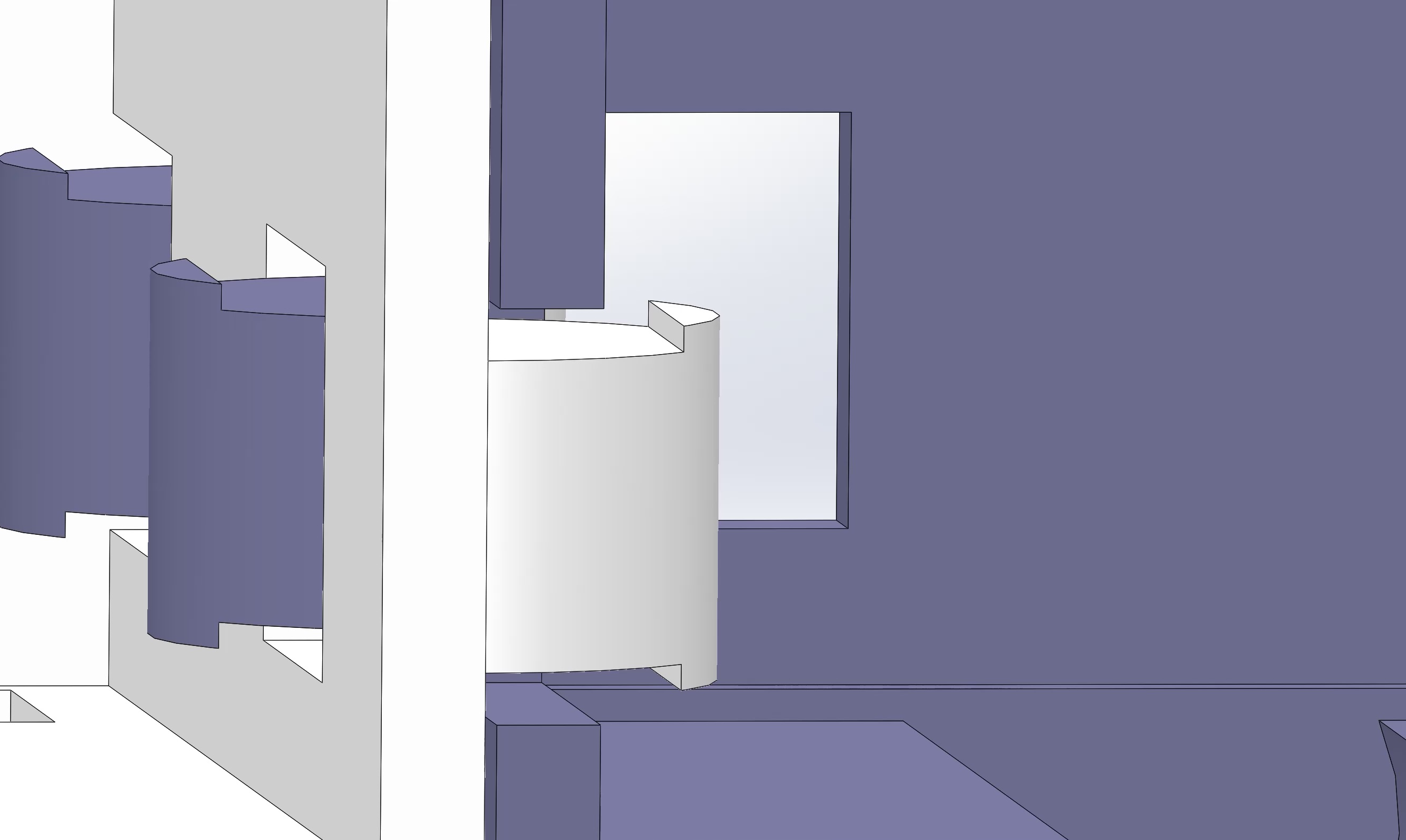}
\caption{}
\label{fig:connect_insert}
\end{subfigure}\hfill
\begin{subfigure}{0.31\textwidth}
\includegraphics[width=\textwidth]{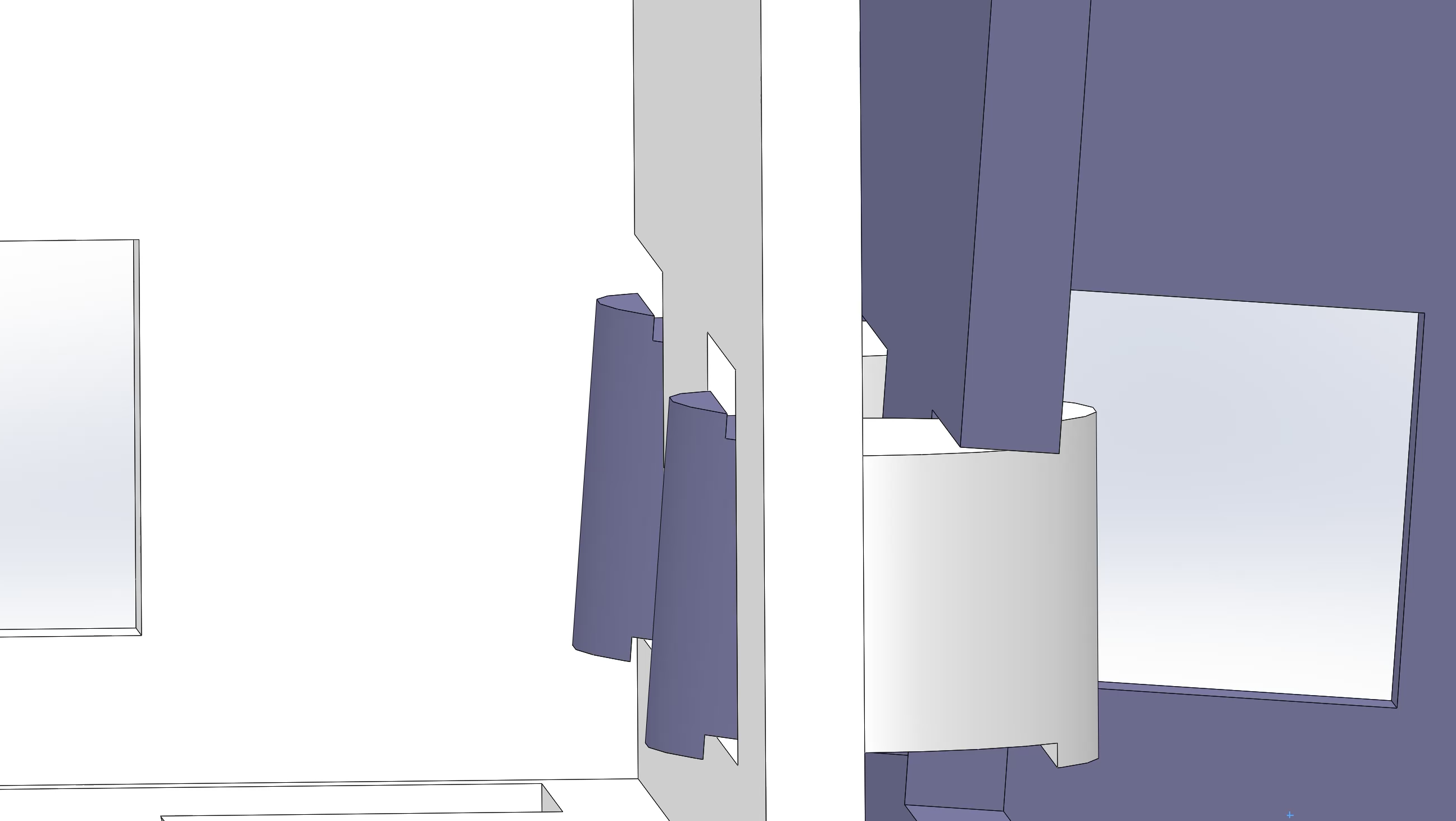}
\caption{}
\label{fig:connect_block}
\end{subfigure}
\setlength{\belowcaptionskip}{-10pt}
\caption{Coupling mechanism between two robots (white and purple). (a) The initial state where two robots are separated from each other. (b) The purple robot can insert its knobs into the holes of the white robot without additional forces. (c) When the purple robot tilts with gravity when it is coming to a gap, the coupling mechanism is activated with hooks on the knobs blocking the movement.}
\label{fig:coupling_mechanism}
\end{figure*}

The body of each robot is 3D printed with NinjaFlex Cheetah TPU with Tensile Modulus 26Mpa. The knobs and holes shown in Figure~\ref{fig:assem_top45} are printed with the body as a single structure. As shown in Figure~\ref{fig:coupling_mechanism}, the coupling mechanism works as the knobs on one piece can fit in the hole of the other puzzle piece. Similarly, the knobs on the robot body are designed to fit in the hole of other robots. As shown in Figure~\ref{fig:connect_not_insert}, initially, the two robots are separated from each other. The figures are zoomed in to focus
on the coupling mechanism for clarification. We are assuming that the two robots are on a flat surface so that the knobs and holes are aligned in the vertical direction. Two robots can move towards each other, as shown in Figure~\ref{fig:connect_insert}. Since the maximum height of the knob, including the top and bottom hooks, is less than the height of the hole, ideally, one robot can slide its knobs into the other robot without any additional force. During manufacturing, the flexible TPU material is used to provide a tight fit between the holes and knobs. As robots move towards a gap, the robot that first leaves the platform will tilt due to gravitational force, resulting in the configuration shown in Figure~\ref{fig:connect_block}. Both friction and the top and bottom hooks will block the movement of the robot falling down the gap. This enables the robots to keep moving to cross the gap towards the other platform, forming a bridge during the process.

The coupling process itself, where knobs are inserted into the holes, does not consume any extra energy other than the energy needed for actuating the robot movement. With this passive coupling mechanism, we can minimize the energy used for coupling compared with other active methods to maximize individual task performances. The connection, once the knobs are fully inserted, are able to hold the weight of 389 robots. The characterization of the knobs used for coupling will be discussed in Section~\ref{sec:charctz_knobs}.

\subsection{Electronics}\label{sec:electronics}
\begin{figure}
\centering
\begin{subfigure}{0.241\textwidth}
\centering
\includegraphics[width=\textwidth]{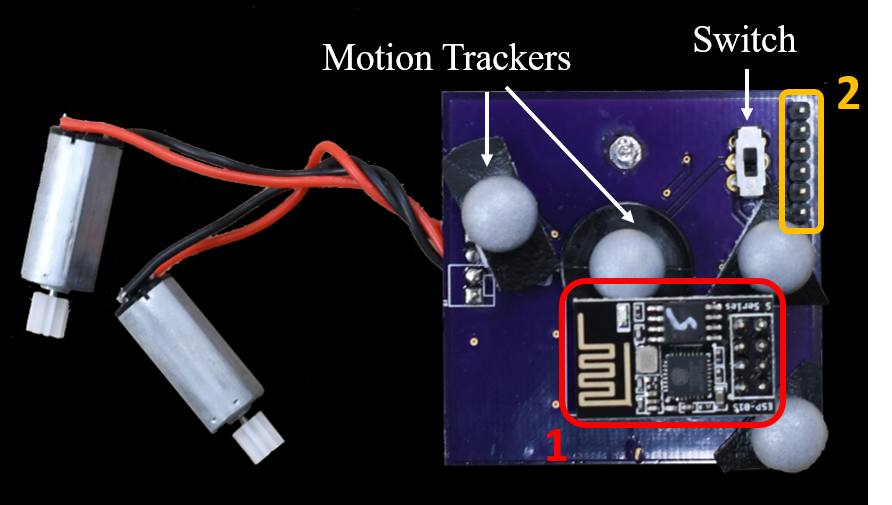}
\caption{}
\label{fig:electronics_top}
\end{subfigure}\hfill
\begin{subfigure}{0.241\textwidth}
\centering
\includegraphics[width=\textwidth]{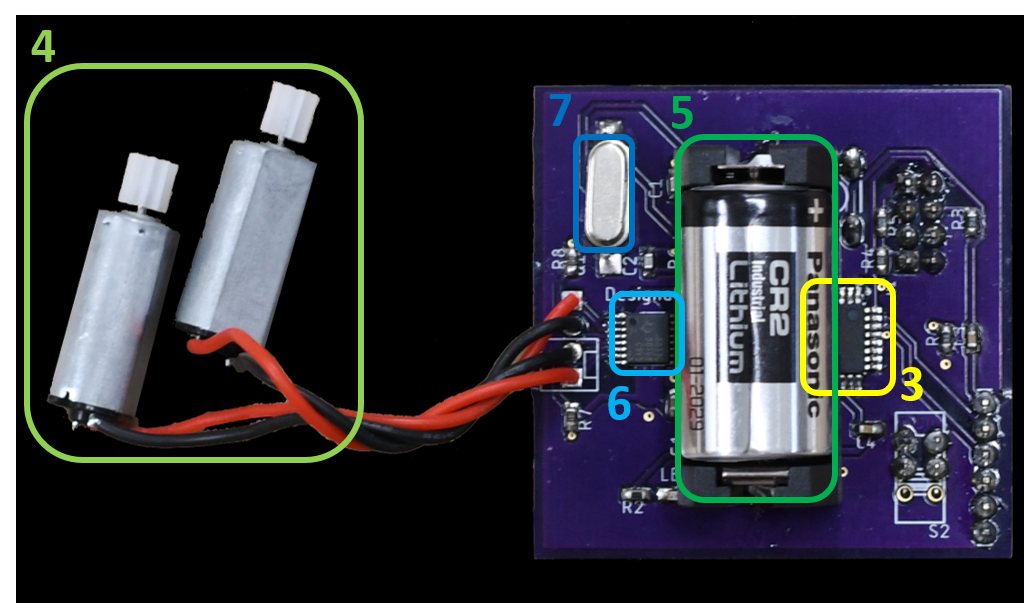}
\caption{}
\label{fig:electronics_bottom}
\end{subfigure}
\setlength{\belowcaptionskip}{-10pt}
\caption{(a) Top view of the circuit board: 1) WiFi module ESP8266, 2) Programming pins; (b) Bottom view of the circuit board: 3) Microcontroller unit ATMega328P, 4) DC Motors, 5) CR2 battery, 6) DRV8833 dual H-bridge motor driver, 7) 8M oscillator.}
\label{fig:electronics}
\end{figure}

The electronics design accommodates the requirements introduced in Section~\ref{sec:robot_design} of on-board power, actuation, communication, and computation. As shown in Figure~\ref{fig:electronics}, the circuit board is printed double-sided with a WiFi module, on-off switch, programming pins on the top, and all other components on the bottom. The whole circuit is powered by a 3V CR2 battery, with a capacity of 850mAh. The battery is available in both rechargeable and non-rechargeable versions. We are using the non-rechargeable ones in the paper for simplicity. The microcontroller unit (MCU) is an 8-bits ATMega328P that operates between 1.8V to 5.5V with an 8M oscillator. It consists of six Pulse-width modulation (PWM) channels, which is convenient to control the motors. The robot communicates with an external computer via the ESP8266 WiFi module. ESP8266 is a low-cost, open-source, small ($14.4 \times 24.7$ mm) WiFi module that supports standard TCP/IP protocol and 2.4GHz WiFi connection while consumes 215 mA current with maximum usage. The integrated AT command interface enables easy communication with the microcontroller. Officially it operates at 3.3V, however with our experiments, it is able to operate normally with voltage as low as 2.8V, thus fitting in our system with the 3V battery supply. The DC motors operate at a voltage from 1V to 4V. These motors can provide a torque of 0.9 mNm with 370 mA current. We can control the velocity of the motor via PWM. The DRV8833 dual H-bridge is a motor driver that has four PWM inputs and four outputs. Each of the PWM inputs is connected with the corresponding MCU PWM output pins, and the four outputs are connected with the two motors (two inputs on each motor). This enables us to control the direction of the current going through the motor, enabling the motors to rotate forward and backward. The DRV8833 operates between 2.7V to 10.8V, with a peak current of 1A per H-bridge.

\subsection{Mechanical Structures and Controls}\label{sec:mech_and_controls}
We are able to control the velocity of the left and right sets of gears by providing pulse-width modulation (PWM) signals to the two motors. In this section, we will present the methods of controlling the velocity of the robots within our designed mechanical structures.

\begin{figure}
    \centering
    \includegraphics[width=0.25\textwidth]{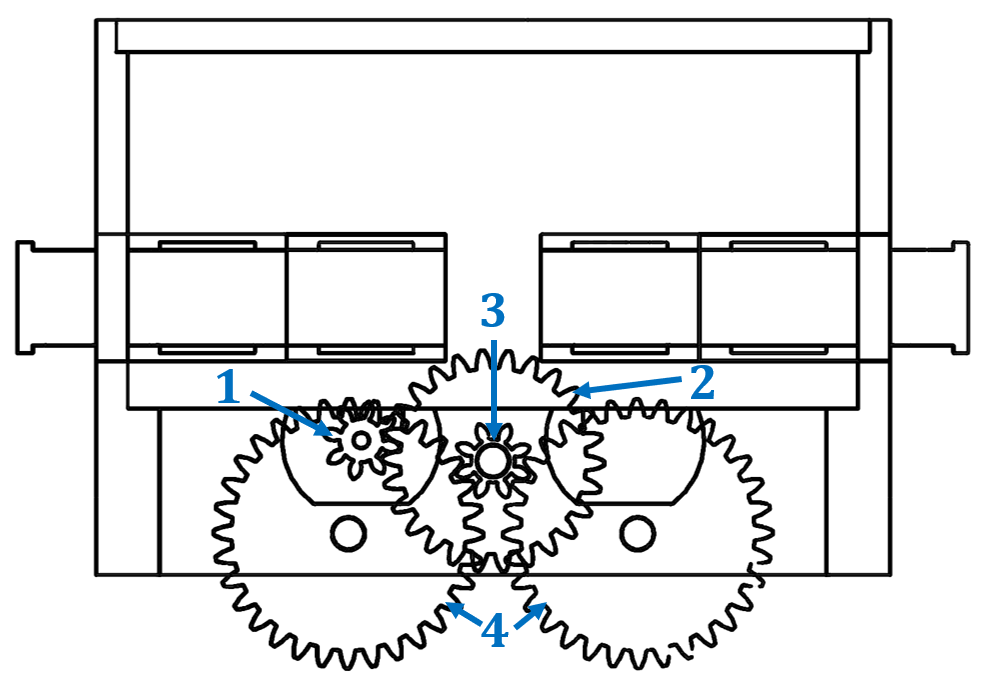}
    \caption{Side view of the robot with double reduction gear. The first pair consists of gear $g_1$ and gear $g_2$ where gear $g_1$ is attached to the motor and $g_2$ is on the center rod. The second part of the double reduction gear consists of $g_3$ and $g_4$, where $g_3$ is attached to $g_2$.
    The two identical side gears are both referred to as $g_4$. The two $g_4$ also serve as wheels. }
    \label{fig:side_gears}
    \vspace{-10px}
\end{figure}

As shown in Figure~\ref{fig:side_gears}, each motor controls one side of the gear sets. The gear $g_1$ is attached to the motor. The gears $g_2$ and $g_3$ form a set of double reduction gears.
The two side gears are identical, both referred to as $g_4$; they also serve as wheels for the robot. This enables larger friction between the ground, and also simplifies the design. We denote the linear velocity of gear $g_i$, $i=\{1,2,3,4\}$, as $v_i$, angular velocity as $\omega_i$, the number of teeth as $z_i$, and reference diameter as $d_i$. All gears have the same module coefficient $M = \frac{d_i}{z_i}$. With the configuration in Figure~\ref{fig:side_gears}, we have $v_1 = v_2,\ \omega_2 = \omega_3,\ v_3 = v_4$, and $\frac{v_i}{\omega_i} = \frac{d_i}{2}$. Since PWM signal controls the angular velocity of the motor $\omega_1$, $v_4 = \omega_1 M \frac{z_1 z_3}{2 z_2}$.

Two side gears on each side are identical, and each side is controlled by an individual motor. For simplicity, we model our robot as a differential drive model. Recall that in Figure~\ref{fig:robot_body}, the forward direction of the robot is aligned with $x$ axis. Thus, we provide forward velocity $v_x$ and angular velocity $\omega$ via WiFi to the robot. With differential drive model, we are able to calculate the velocity needed on the left $v_l$ and right $v_r$ as \cite{lavalle2006planning}
\begin{equation}
v_r = \frac{2v_x + \omega L}{2R},\ v_l = \frac{2v_x - \omega L}{2R} \label{eq:diff_drive}
\end{equation}
where $L$ is the length between the wheels and $R$ is the radius of the wheels, i.e. gear $g_4$. The output PWM signal from the MCU controls the rotational speed of each individual motor denoting as $\omega_r$ on the right and $\omega_l$ on the left. By combining the equations of the gear sets and the differential drive, we have
\begin{equation}
\omega_r = \frac{z_2(2v_x + \omega L)}{M z_1 z_3 R},\ \omega_l = \frac{z_2(2v_x - \omega L)}{M z_1 z_3 R}
\end{equation}
In our design, we have $z_1 = 8,\ z_2 = 26,\ z_3=8,\ M = 0.5,\ L = 40\ mm, R = 17\ mm$. During implementation, by substituting these values, we are able to control the robot accordingly.

\subsection{Swarms Coupling Behaviors}
The algorithms used for the robots to couple and decouple are one-dimensional rendezvous and anti-rendezvous behaviors for swarms. Rendezvous swarms behavior is a consensus algorithm where each robot communicates with its neighbors to move towards a direction that will eventually gather everyone together \cite{olfati2004consensus, olfati2007consensus}. Consider our system of $N$ robots on a one-dimensional line, we denote the position of robot $i$ as $x_i\in \mathbb{R}$ with control input $\dot x_i = u_i$, where $i=\{1, \dots, N\}$. In our system setup, all robots are able to communicate with each other via a central computer. Therefore, we can simplify the rendezvous controller as
\begin{equation}
\dot x_i = \frac{1}{N}\sum_{j \neq i}(x_j - x_i)
\end{equation}
As a result, the robots will move towards each other until they are physically coupled. Once they are successfully coupled ($\min||x_i - x_j|| = \text{robot body length}, \forall i,j\in{1, \dots, N}, i\neq j$), they can perform other behaviors as one connected component.

Similarly, the robots are able to decouple with each other via 1D anti-rendezvous:
\begin{equation}
\dot x_i = -\frac{1}{N}\sum_{j \neq i}(x_j - x_i)
\end{equation}
As a result, the robots will move to decouple with each other and perform individual tasks later on.

However, due to actuation uncertainty, the robots might not stay precisely aligned during this 1D rendezvous behavior. In actual experiments, we utilize the environment to reduce these uncertainties, e.g., having one robot stay against a wall. 

\section{Experiments}\label{sec:experiments}
We characterized the coupling knobs for maximum performance of the coupling mechanism. We then performed experiments of up to nine robots for the gap-crossing behavior with different environmental parameters. We also present frames from the video where robots couple, cross a gap, decouple, and visit individual goals.

\subsection{Characterizing the Coupling Knobs}\label{sec:charctz_knobs}
The knobs and holes are the core part of the coupling mechanism, and the dimensions determine the performance of connections. It is ideal to have the height of the knobs (including the hooks) to match precisely the height of the holes. However, 3D printed surfaces, especially surfaces that need support underneath, are generally not smooth. Although TPU is a flexible material, having a tight fit will require more torque from the actuators. Thus, by experimenting with different parameters, we design the height of the knobs to be 1mm less than that of the holes. 

\begin{figure}
\centering
\begin{subfigure}{0.49\columnwidth}
\centering
\includegraphics[width=\textwidth]{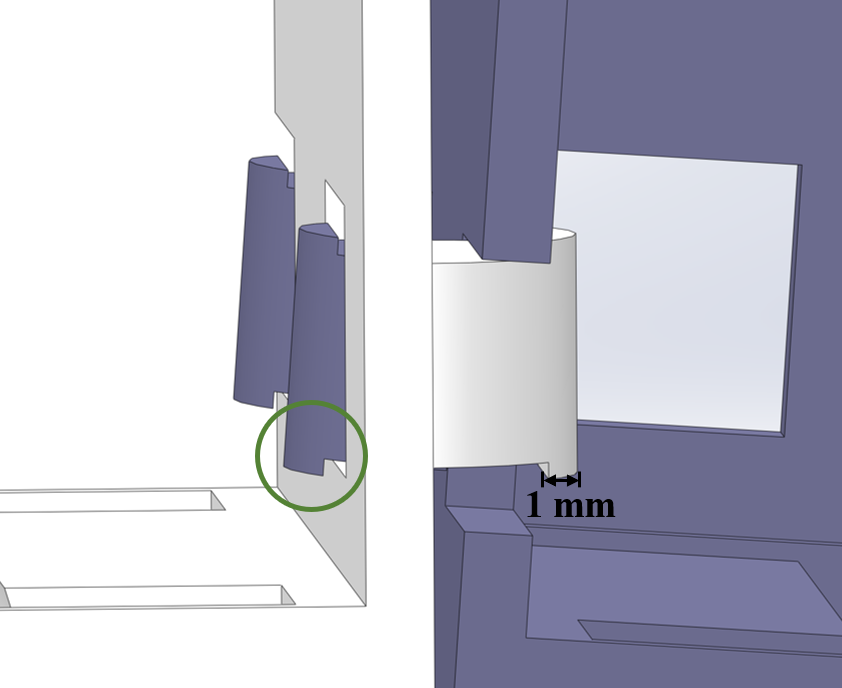}
\caption{}
\label{fig:success_lock_1mm}
\end{subfigure}
\begin{subfigure}{0.49\columnwidth}
\centering
\includegraphics[width=\textwidth]{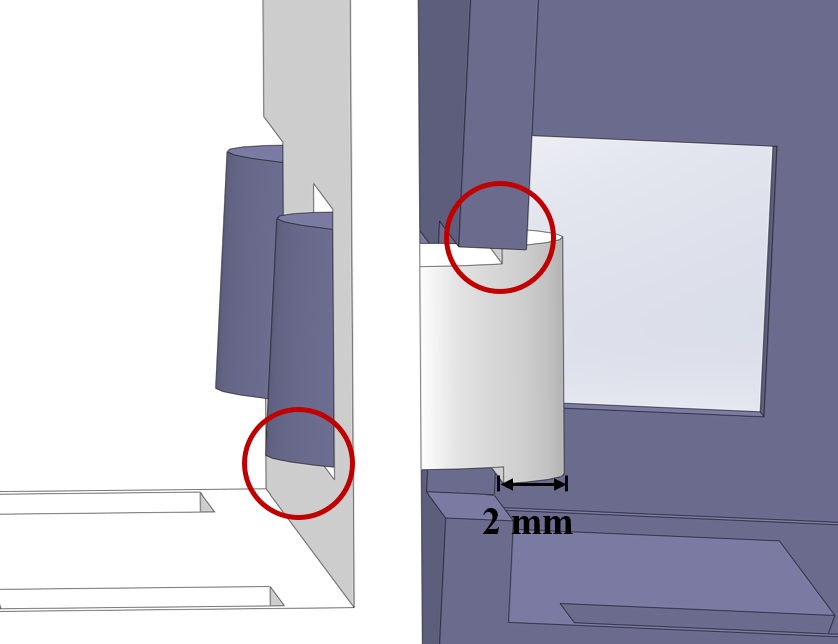}
\caption{}
\label{fig:not_success_lock_2mm}
\end{subfigure}
\caption{(a) A successful coupling example with hook width 1mm when the robot is tilted due to gravity. The hooks are able to block the movement of the robot body. (b) An unsuccessful coupling example with hook width 2 mm. Since the hook width is too large, it relies only on friction between surfaces, not the hooks, to block the movement of the robot body. }
\label{fig:success_locks}
\end{figure}

The size of the hook on the knob is also an essential parameter for coupling. Two possible coupling status with different hook width is shown in Figure~\ref{fig:success_locks}. In Figure~\ref{fig:success_lock_1mm}, the hook width is 1 mm and the hooks can lock the movement of the robot on the right when it is tilted. When the robots are moving towards the gap, the larger the tilting angle, the lower the front of the robot will be, and the smaller the gap they will be able to cross. The larger the hook width gives a smaller tilting angle, and thus better gap-crossing performance. However, as shown in Figure~\ref{fig:not_success_lock_2mm}, when the hook width is too large (still smaller than \textit{knob radius} $-$ \textit{wall thickness}), the hooks may fail to lock the movement. 

\begin{figure}
\centering
\includegraphics[width=0.7\columnwidth]{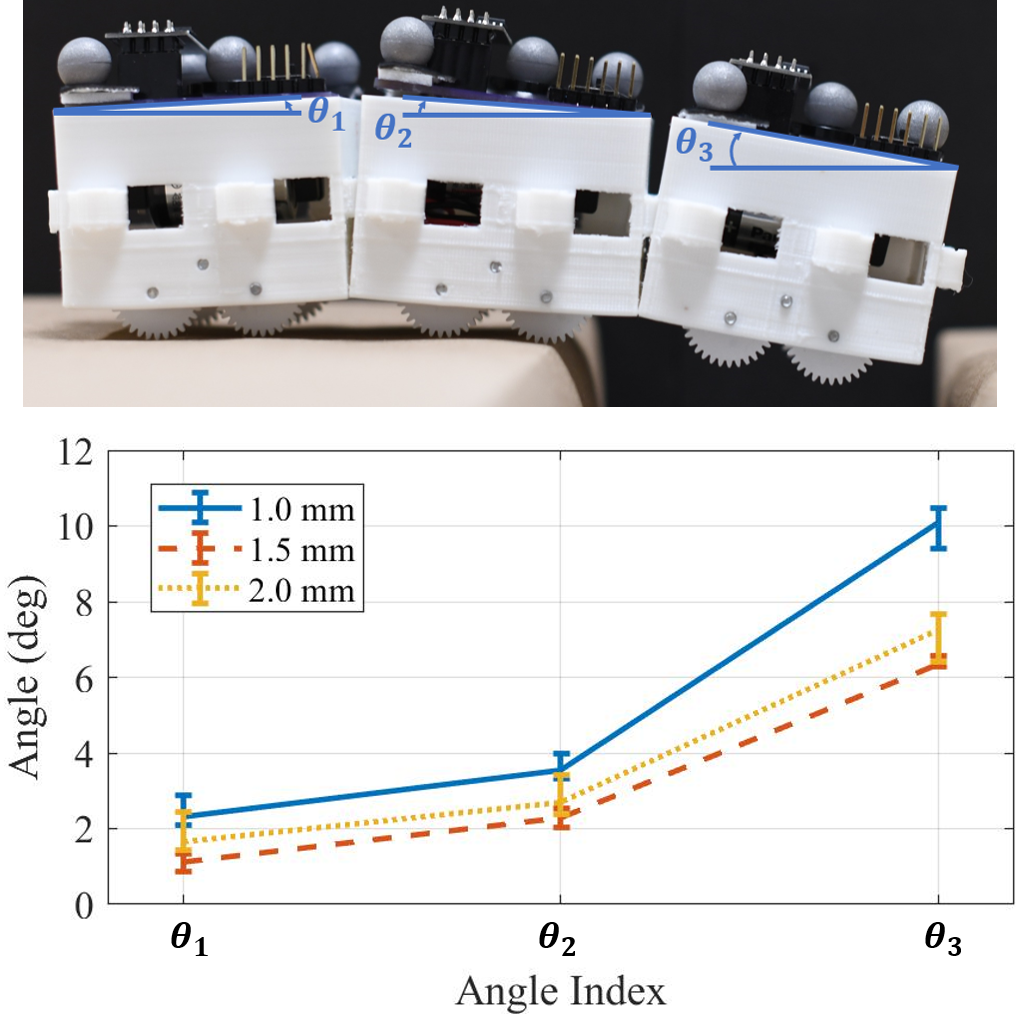}
\setlength{\belowcaptionskip}{-10pt}
\caption{The tilting angles of robots, with different sets of hook width (1.0, 1.5, 2.0 mm) when they move towards a gap. The measurement is performed when the the whole assembly is just about to move off the original platform.}
\label{fig:angle_result}
\end{figure}

As shown in Figure~\ref{fig:angle_result}, we measure the tilting angles of robots when they move towards a gap. The experiments are done with three robots, and the measurement is performed when the second robot is just about to leave the platform. This is the moment when the tilting angles are the largest. Due to gravity and the coupling, the remaining two will also tilt when the first robot tilts. We measure their angles $\theta_1$, $\theta_2$, $\theta_3$ with respect to the horizontal plane. The 1 mm hook width gives the largest angle, while the 1.5 mm hook width gives the smallest tilting angle. Therefore, in the remaining experiments, all robot knobs have a 1.5 mm hook for maximum performance. 

\subsection{System Experiments}
\subsubsection{Gap-crossing Performance}

\begin{figure*}[tbh]
\centering
\begin{subfigure}{0.32\textwidth}
\centering
\includegraphics[width=\textwidth]{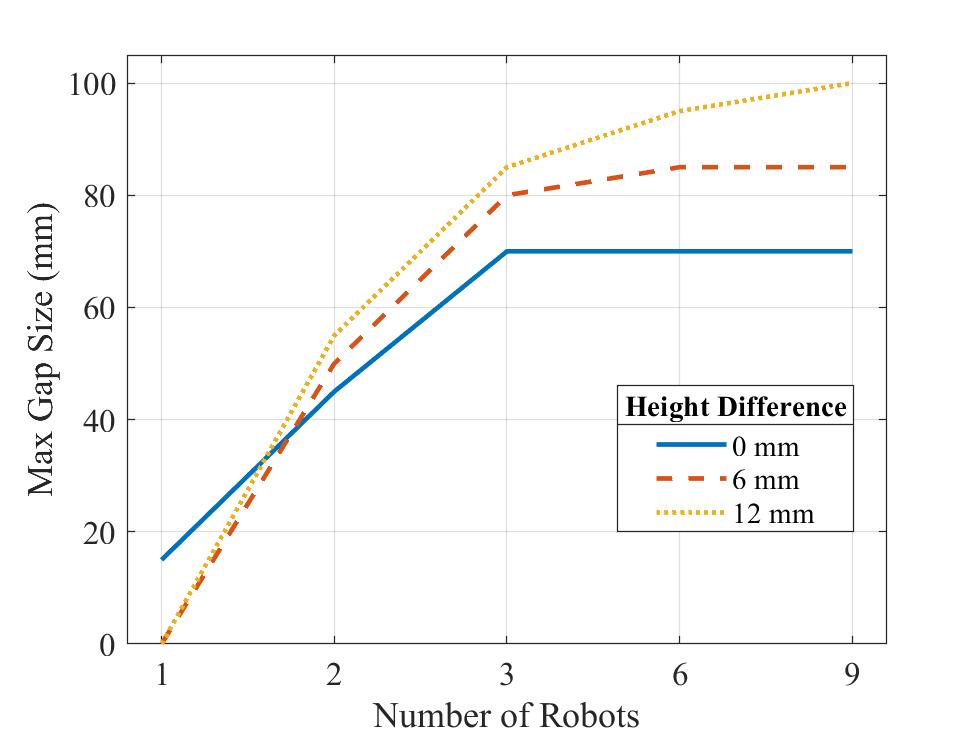}
\caption{}\label{fig:max_gap_vs_rnum}
\end{subfigure}
\begin{subfigure}{0.32\textwidth}
\centering
\includegraphics[width=\textwidth]{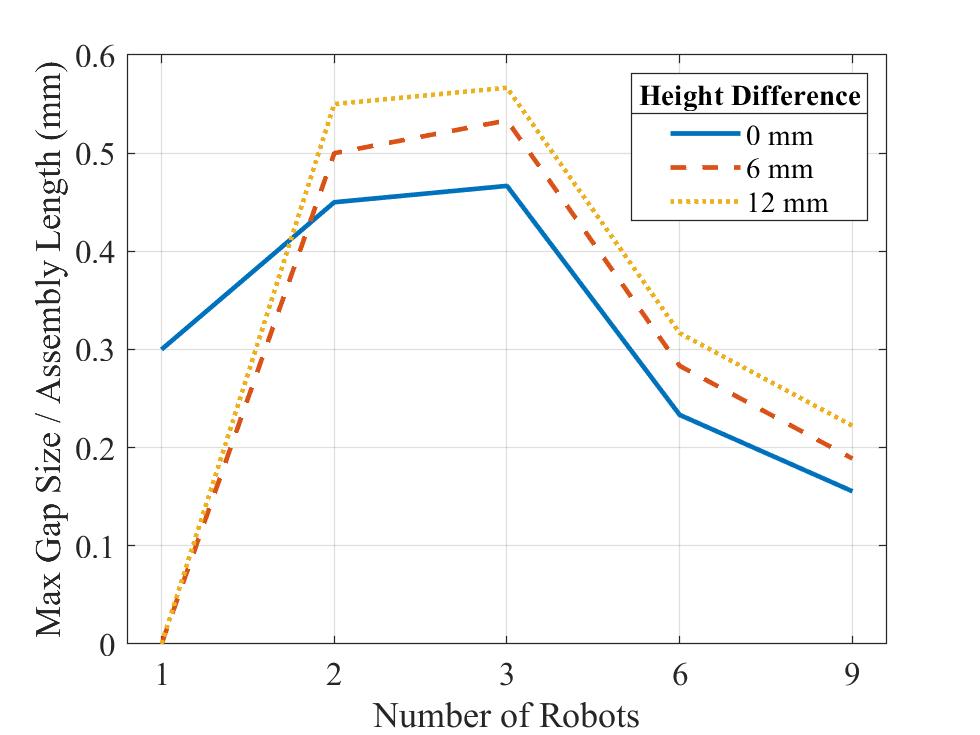}
\caption{}\label{fig:max_gap_assem_len_vs_rnum}
\end{subfigure}
\begin{subfigure}{0.32\textwidth}
\centering
\includegraphics[width=\textwidth]{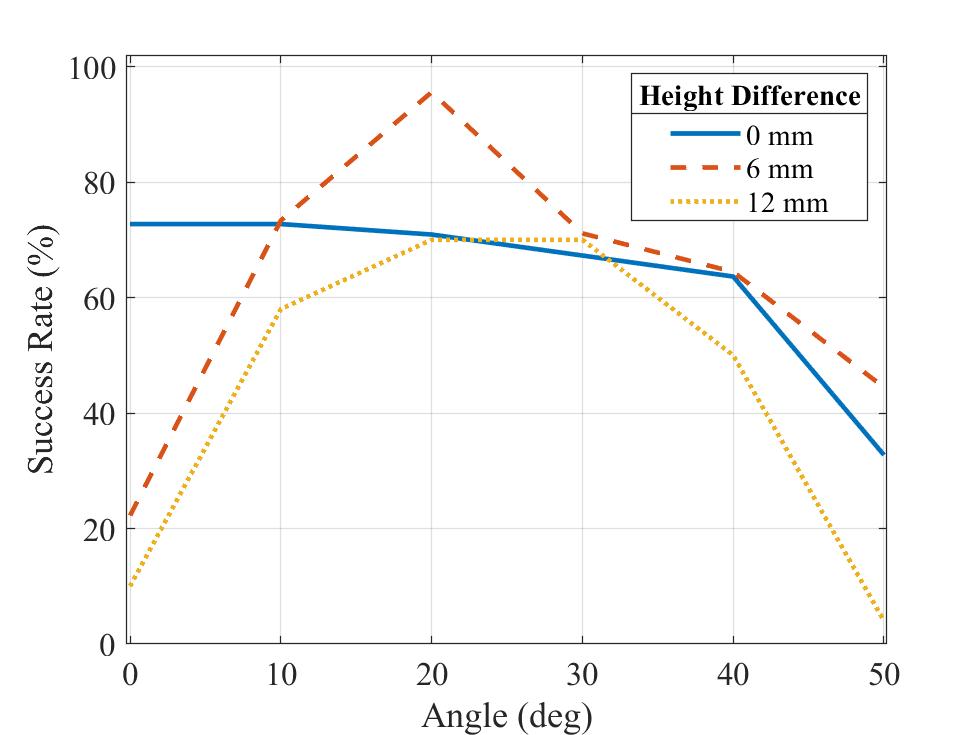}
\caption{}\label{fig:success_vs_angle}
\end{subfigure}
\caption{(a) Maximum gap size that the robots are able to cross (regardless of heading angle) versus the number of robots. (b) Ratio of the maximum gap size with respect to the length of the whole assembly (\textit{robot number} $\times$ \textit{body length}) versus the number of robots. (c) Success rate of all experiments under the same heading angle.}
\label{fig:result_plots}
\end{figure*}

The length of each robot is 50 mm. We analyze the performances of the gap-crossing behavior with different variables: number of robots (1, 2, 3, 6, 9), length of the gap (10 mm to 100 mm), heading angle ($0^{\circ}$ to $50^{\circ}$), height difference between the two platforms (0 mm, 6 mm, 12 mm, the starting platform is higher than the target platform). We perform five runs for each combination and recorded the number of successful runs, i.e. all of the robots crossed the gap while staying coupled. 

In Figure~\ref{fig:result_plots}, we present the result of the maximum gap sizes different number of robots can cross (\ref{fig:max_gap_vs_rnum}), the ratio of maximum gap size with the length of the whole robot assembly (\ref{fig:max_gap_assem_len_vs_rnum}), and the relation between success rate and the heading angle of the robots (\ref{fig:success_vs_angle}). In Figure~\ref{fig:max_gap_vs_rnum}, we consider a gap size and height difference that the robots can cross when the success rate is higher than 50\%. We can see that as the number of robots increases, they can cross over a larger gap. However, with a gap between the platforms of the same height, increasing the number of robots does not increase performance. The major bottleneck is with the tilting angle mentioned in Section~\ref{sec:charctz_knobs}. The tilting angle does not increase or decrease with the change of robot number. This bottleneck persists with a larger height difference, but the height difference in platforms compensate for the height drop in the robot assembly. This results in better performances as the robot number increases with larger platform height difference. In Figure~\ref{fig:max_gap_assem_len_vs_rnum}, the ratio of the maximum gap size and the length of the assembly shows the effectiveness of increasing the number of robots. However, although the robot assembly can cross larger gaps with more robots, i.e., longer assembly length, the significance of increasing the number of robots decreases after the three robots setting. In our experiments with different heading angles, the edge of each platform aligns with the y-axis of the world frame. Thus, the robot with a heading angle of $0^\circ$ is perpendicular to the platform edge. Figure~\ref{fig:success_vs_angle} shows the success rate of all number of robots, given the specific heading angles. Failure cases include 1) robots fail to proceed once reaching the opposite platform due to tilting angle; 2) connections broke due to impact when robots reach the other platform; 3) robot tumbled when reaching the other platform (only for single robot case). With the same height platforms, the success rate gradually drops as the angle increases because of the traversing distance across the gap increases. However, with a larger height difference, angles around $20^\circ$ give better performances. Since the whole robot assembly will only fall off when the center of mass of the assembly leaves the original platform, having an angle with the platform edge will increase the length protruding the platform. This enables the diagonal of the robots to reach the other side first. This did not give better performances with platforms of the same height because of the height drop of the robots due to the tilting angle.

\subsubsection{Combined Behaviors}

\begin{figure}
\centering
\includegraphics[width=0.8\columnwidth]{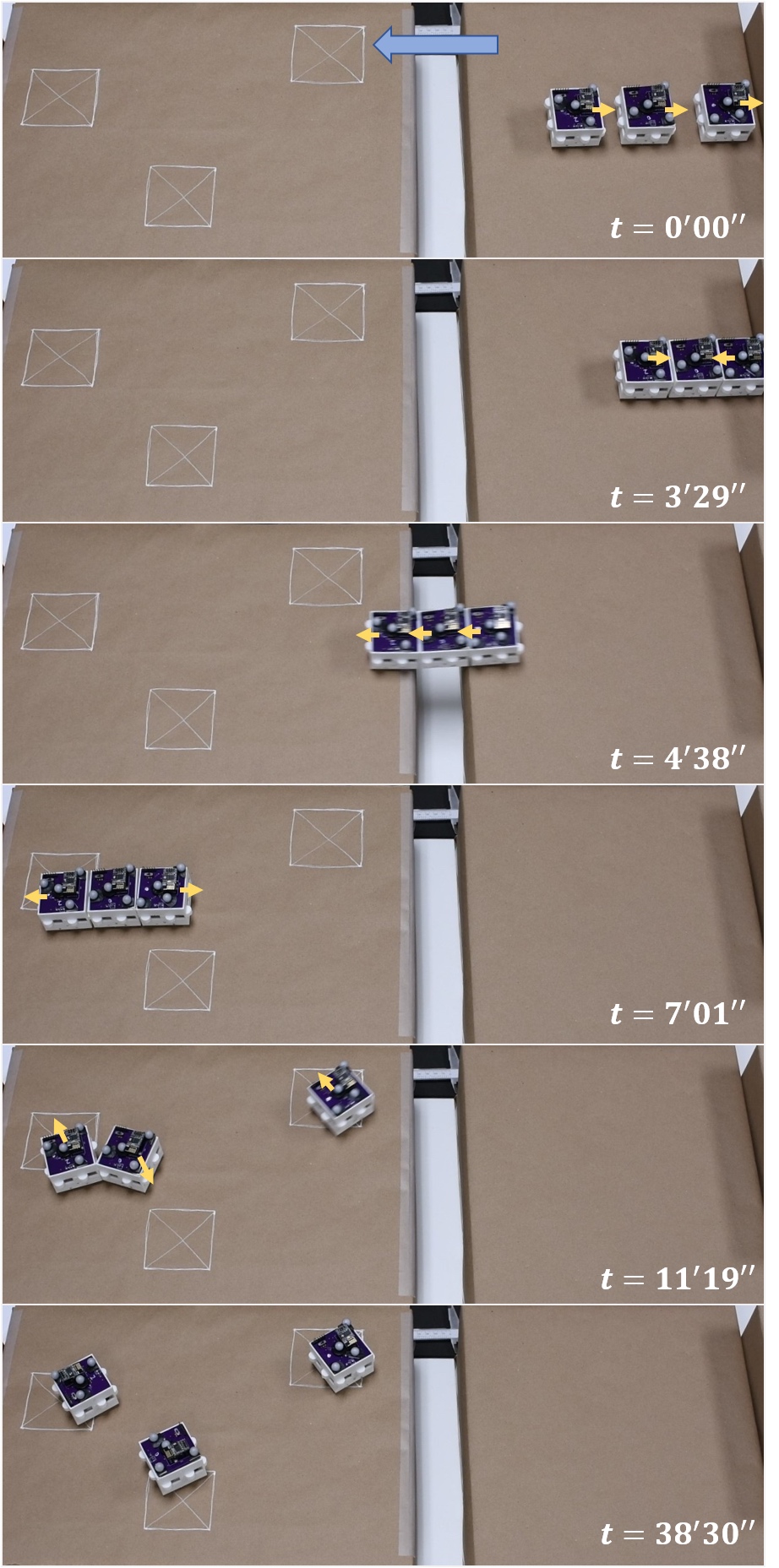}
\caption{These video frame sequences show that three initially separated robots are able to couple with each other, go cross a gap (60 mm), decouple, and go to their individual goal locations (white squares). In the first frame, the robots start on platform on the right. They will need to cross the gap and reach the left platform. The blue arrow shows their goal direction, and yellow arrows show their current direction of motion of each robot. }
\label{fig:cross_screenshots}
\end{figure}

To demonstrate the ability of the PuzzleBots to assembly and disassembly autonomously, we present a sequence of video frames of our hardware system, as shown in Figure~\ref{fig:cross_screenshots}. The original video is included as a supplement. Three robots initially separated are located on the right platform. The right platform is 6 mm higher than the left. The gap between the two platforms is 60 mm wide. There is a wall on the right, aligned with the platform on the right. The left and middle robots run the 1D rendezvous controller, as described in Section~\ref{sec:mech_and_controls}, while the right robot runs into the wall. The robots can couple with each other against the wall. As soon as the minimum distance between robots reaches their body length, they will move towards the gap. With the coupling mechanism, they can cross the gap and reach the other platform. All three run the anti-rendezvous controller when they have successfully crossed the gap. Once they decouple with each other, the robots will move towards their individual goal locations.

\section{Conclusion and Discussion}\label{sec:conclusion}
In this paper, we have introduced the PuzzleBots, a robotic swarm system where robots can couple with each other to form functional structures without additional energy for coupling, while maintaining individual mobility for completing different tasks. We utilize knobs and holes on the robot body to perform the coupling mechanism. We show with hardware experiments that the robots can cross gaps approximately half the size of the whole assembly and can couple and decouple autonomously based on task requirements.

While we show that the robots can couple in the front and back, the first step of our future work tries to realize horizontal coupling mechanism. Robots coupling in the left and right to form a mesh-like structure may further extend the performance of the gap-crossing behavior. Although we have trials that show the possibility of pushing into each other from the side to couple, controlling this behavior is currently under investigation. Note that this is the first version of Puzzlebots. Future development may include on-board sensors, rechargeable batteries, improved wheel design, and a further decentralized system without a central computer to bring the system to real-world applications.

Furthermore, our work utilizes the environment to reduce actuation uncertainty during execution. However, a systematic way of when and how to best make use of the environment remains an open question. Additionally, we are also interested in the 3D coupling, where more flexibility can be introduced to form 3D structures like ropes or nets. We hope this contribution can provide benefits to robotic applications in uncertain and complex environments.

\bibliographystyle{plain}
\bibliography{ref}

\end{document}